\documentclass[11pt]{article}

\usepackage[final]{acl}

\usepackage{times}
\usepackage{latexsym}

\usepackage[T1]{fontenc}

\usepackage[utf8]{inputenc}

\usepackage{microtype}

\usepackage{inconsolata}

\usepackage{graphicx}

\usepackage{amsmath,amssymb,amsfonts,mathtools}
\usepackage{amsthm}
\usepackage{bbm}
\usepackage{geometry}

\title{QUBO-Optimized Evidence Selection for Retrieval-Augmented Question Answering with Unconventional Solvers}

\author{Rahul Singh\\  University of California Santa Barbara \\   Santa Barbara, California \\  \texttt{rps@ucsb.edu} \\ 
\And
Madhav Vadlamani\thanks{Equal Contribution} \\ Georgia Institute of Technology\\ Atlanta, Gerogia\\ \texttt{mvadlamani6@gatech.edu}}

\begin{document}
\maketitle
\begin{abstract}
Retrieval-augmented question answering depends on selecting an appropriate set of evidence passages before answer generation. However, many retrieval pipelines rely on top-\(k\) ranking, where passages are selected primarily by individual relevance scores, even though multi-hop questions often require passages that jointly satisfy multiple information requirements. Recent LLM-based methods address this limitation by treating retrieval as a set-selection problem, but using an LLM for this intermediate selection stage can be expensive and difficult to scale. In this work, we formulate evidence selection as a Quadratic Unconstrained Binary Optimization (QUBO) problem. Given a question, candidate passages, and decomposed information requirements, our method constructs an energy function that balances passage relevance, requirement coverage, support strength, redundancy, complementarity, and compactness. Low-energy configurations correspond to evidence subsets that cover the necessary requirements while avoiding unnecessary or repetitive context. The selected passages are then passed to a downstream language model for answer generation, separating the combinatorial evidence-selection step from the semantic answer-generation step. We evaluate the proposed QUBO selector on HotpotQA, a multi-hop question-answering benchmark, and compare it with LLM-based set selection as well as non-LLM baselines including BM25, relevance top-\(k\), maximal marginal relevance, hybrid lexical-semantic ranking, greedy coverage, and random selection. The QUBO selector achieves competitive exact-match and token-F1 performance relative to LLM-based set selectors, while providing a solver-compatible formulation for structured evidence selection. These results suggest that multi-hop evidence selection can be cast as a discrete optimization problem, opening a path toward retrieval pipelines in which LLMs are reserved primarily for answer generation while the selection stage is executed by unconventional Ising machines, quantum annealers, quantum-inspired optimizers, or gate-model quantum algorithms for QUBO solving.
\end{abstract}

\section{Introduction}

Large language models (LLMs) have demonstrated strong capabilities in natural language understanding and generation, but their responses remain constrained by both the information encoded in their parameters and the quality of the context provided at inference time. Retrieval-Augmented Generation (RAG) addresses this limitation by coupling a parametric language model with an external non-parametric knowledge source, allowing generation to be conditioned on retrieved evidence rather than relying solely on memorized knowledge~\cite{Lewis2020}. This paradigm is especially useful for knowledge-intensive tasks requiring factual grounding, provenance, domain adaptation, and access to dynamically updated information~\cite{Gao2024,Barnett2024}. However, RAG performance is highly sensitive to the selected context: retrieval errors, missing evidence, redundant passages, and irrelevant distractors can degrade generation, especially in multi-hop question answering where the answer may require combining complementary evidence across passages~\cite{Yang2018,Yan2024}. Simply increasing the number of retrieved passages is also insufficient, since long-context language models may fail to use relevant evidence depending on its position and salience~\cite{Liu2024}.

Most RAG systems therefore follow a retrieve-then-rerank pipeline, where a sparse, dense, or hybrid retriever produces candidate passages, a reranker orders them, and the top-\(k\) passages are passed to the generator. While cross-encoder and LLM-based listwise rerankers improve ranking quality~\cite{Sun2023,Liu2025}, they typically optimize passage ordering rather than the final evidence subset directly. This distinction matters because set-level properties such as requirement coverage, redundancy, complementarity, and joint answer support are not always explicitly optimized~\cite{Lee2025}. Diversity-aware methods such as maximal marginal relevance partially address redundancy by balancing relevance and novelty~\cite{Carbonell1998}, but they do not directly model query-specific information requirements or optimize requirement-level coverage within a unified subset-selection objective. At the same time, large-scale retrieval and reranking can introduce substantial deployment cost, latency, memory-capacity, and data-movement challenges~\cite{hsu2026}. These observations motivate a set-selection view of RAG context construction~\cite{Lee2025,jiang2026}, where the goal is not merely to select the highest-ranked individual passages, but to construct a compact evidence subset that collectively satisfies the informational requirements of the query.

This set-selection perspective also creates an opportunity for specialized optimization hardware. Unlike LLM-based reranking or prompting, which requires repeated transformer inference, a QUBO/Ising formulation exposes the context selection decision as an explicit binary energy minimization problem. Such problems can be mapped to quantum annealers and gate-based quantum optimization algorithms, and can also be executed by quantum-inspired hardware such as digital annealers, coherent Ising machines, and CMOS-compatible Ising machines.These platforms are designed to search the energy landscape for low energy configurations using physical dynamics or hardware-specific optimization rules. Prior work has reported problem dependent speedups for digital annealers on fully connected QUBO/spin-glass instances, scalable coherent-Ising-machine behavior on large graph-optimization problems, and improved execution-time or energy characteristics for CMOS-compatible Ising designs~\cite{Aramon2019,Haribara2016,Zhang2022}. While we do not assume a universal quantum speedup, this hardware direction is attractive for large-scale RAG pipelines because the candidate pool and pairwise passage interactions can grow substantially with corpus size, making the final context-selection step a natural target for low-latency and potentially energy-efficient QUBO/Ising acceleration.

In this work, we formulate requirement-aware passage selection as a Quadratic Unconstrained Binary Optimization (QUBO) problem. QUBO provides a standard framework for binary optimization with linear utility terms and quadratic interaction terms~\cite{Glover2022}, making it well suited to RAG context construction: each candidate passage is represented by a binary variable, while the objective rewards relevance, requirement coverage, support strength, and complementarity, and penalizes redundancy and unnecessary context length. Low-energy solutions therefore correspond to compact evidence subsets selected for their collective utility rather than only their individual relevance scores.

The proposed formulation adds a structured optimization layer between retrieval and answer generation. It does not replace the retriever or the generator; instead, it replaces the intermediate context-selection step with an explicit set-optimization problem over an already retrieved candidate pool. The selected passages are then passed to a downstream language model for answer generation, separating combinatorial evidence selection from semantic answer synthesis. To the best of our knowledge, this is the first work to formulate requirement-aware RAG context selection as a QUBO problem, enabling the selection stage to be mapped to Ising/QUBO-compatible solvers such as simulated annealers, quantum annealers, digital annealers, coherent Ising machines, CMOS/SRAM-based Ising machines, and gate-model quantum optimization algorithms after suitable mapping~\cite{Ceselli2023,Seker2022,Wei2026,vadlamani2026,Perlin2026,He2025}. This solver-compatible representation is particularly relevant as candidate pools and retrieval corpora grow: specialized QUBO and Ising accelerators can target the combinatorial selection step directly, offering a path toward lower-latency and potentially more energy-efficient context selection than repeated LLM-based reranking or set-selection calls.

The main contributions of this paper are summarized as follows:
\begin{itemize}
    \item We formulate RAG context construction as a requirement-aware binary subset-selection problem, where the selected evidence is optimized at the set level rather than obtained through independent top-\(k\) ranking.

    \item We derive a QUBO objective for evidence selection that jointly encodes passage relevance, requirement coverage, support strength, redundancy reduction, complementarity, compactness, and selection-budget constraints within a single solver-compatible energy function.

    \item We present a modular RAG pipeline that separates semantic preprocessing, combinatorial context selection, and answer generation: semantic scores define the QUBO instance, the solver selects the context subset, and a downstream language model generates the final answer.

    \item We evaluate the proposed approach on multi-hop question answering using both downstream answer-generation metrics and set-level evidence-selection metrics, comparing against SetR-style LLM set selectors and non-LLM baselines including BM25, relevance top-\(k\), maximal marginal relevance, hybrid lexical--semantic ranking, greedy coverage, and random selection.
\end{itemize}

\section{Background}
\label{sec:background}

The proposed method modifies the context-selection stage of Retrieval-Augmented Generation (RAG). Rather than reintroducing the full RAG pipeline, this section summarizes the context-selection problem and then introduces Quadratic Unconstrained Binary Optimization (QUBO), which provides the mathematical basis for the proposed formulation.

\subsection{RAG Context Selection}
\label{Sec:RagContextSelection}

RAG augments a parametric language model with retrieved non-parametric evidence, allowing answer generation to be conditioned on external context rather than solely on memorized model parameters~\cite{Lewis2020}. In a standard pipeline, an input query is used to retrieve or rerank a candidate pool of passages, and the final context is commonly formed by selecting the top-\(K\) passages according to a relevance or reranking score. If \(\mathcal{C}_q\) denotes the candidate pool for query \(q\), this selection can be written as
\begin{equation}
    \mathcal{S}_q^{\mathrm{top}\text{-}K} = \arg\max_{\mathcal{S}\subseteq\mathcal{C}_q,\ |\mathcal{S}|=K} \sum_{p_i\in\mathcal{S}}\rho(q,p_i),
\end{equation}
where \(\rho(q,p_i)\) is the score assigned to passage \(p_i\). This additive objective selects passages according to their individual scores, but it does not explicitly optimize interactions among selected passages.

This limitation is important in multi-hop and requirement-sensitive question answering, where the answer may require combining complementary evidence distributed across multiple passages~\cite{Yang2018}. A passage with high individual relevance may repeat evidence already present in another passage, while a moderately ranked passage may be necessary because it supplies a missing bridge fact or completes an evidence chain. Recent work has therefore argued for set-level context construction, where passages are selected according to how well they collectively satisfy the information needs of the query rather than only by their independent relevance scores~\cite{Lee2025,jiang2026}. This motivates treating RAG context construction as a binary subset-selection problem.

\subsection{Quadratic Unconstrained Binary Optimization}
\label{sec:qubo_background}

Quadratic Unconstrained Binary Optimization (QUBO) is a standard formulation for discrete optimization over binary variables~\cite{Glover2022}. Given a binary vector
\begin{equation}
\mathbf{x}
=
[x_1,x_2,\dots,x_n]^{\top},
\qquad
x_i\in\{0,1\},
\end{equation}
a QUBO problem is commonly written as
\begin{equation}
\label{eq:qubo_matrix_form}
\min_{\mathbf{x}\in\{0,1\}^{n}}
E(\mathbf{x})
=
\mathbf{x}^{\top}Q\mathbf{x},
\end{equation}
where \(Q\in\mathbb{R}^{n\times n}\) is the QUBO coefficient matrix. The diagonal entries \(Q_{ii}\) encode linear costs or rewards because \(x_i^2=x_i\) for binary variables, while the off-diagonal entries encode pairwise interactions between variables. Equivalently, using an upper-triangular convention,
\begin{equation}
E(\mathbf{x})
=
\sum_{i=1}^{n} Q_{ii}x_i
+
\sum_{1\le i<j\le n} Q_{ij}x_i x_j .
\end{equation}

QUBO objectives are closely related to Ising model Hamiltonians~\cite{Lucas2014} through the affine transformation
\begin{equation}
s_i = 2x_i - 1,
\qquad
s_i\in\{-1,+1\}.
\end{equation}
This relationship makes QUBO a solver-compatible representation for simulated annealing and for Ising/QUBO-oriented optimization backends, including quantum annealers, digital annealers, coherent Ising machines, CMOS/SRAM-based Ising machines, and gate-model quantum optimization algorithms after suitable mapping~\cite{Ceselli2023,Seker2022,Wei2026,vadlamani2026,Perlin2026,He2025}. In this work, the QUBO representation is used to encode requirement-aware passage selection, and the construction of the corresponding matrix \(Q\) is detailed in Sec.~\ref{sec:methods}.

\section{Methods}
\label{sec:methods}

This section presents the proposed framework for set-based context selection in RAG. The goal is to select a compact subset of passages that provides sufficient evidence for answer generation. In contrast to retrieve-then-rerank pipelines that typically pass the top-\(k\) ranked passages to the generator, the proposed framework treats context construction as a binary subset-selection problem. This allows the final context to be optimized with respect to passage relevance, requirement coverage, support strength, complementarity, redundancy, and compactness.

\subsection{Pipeline Overview}
\label{sec:pipeline_overview}

\begin{figure*}[t]
\centering
\includegraphics[width=\linewidth]{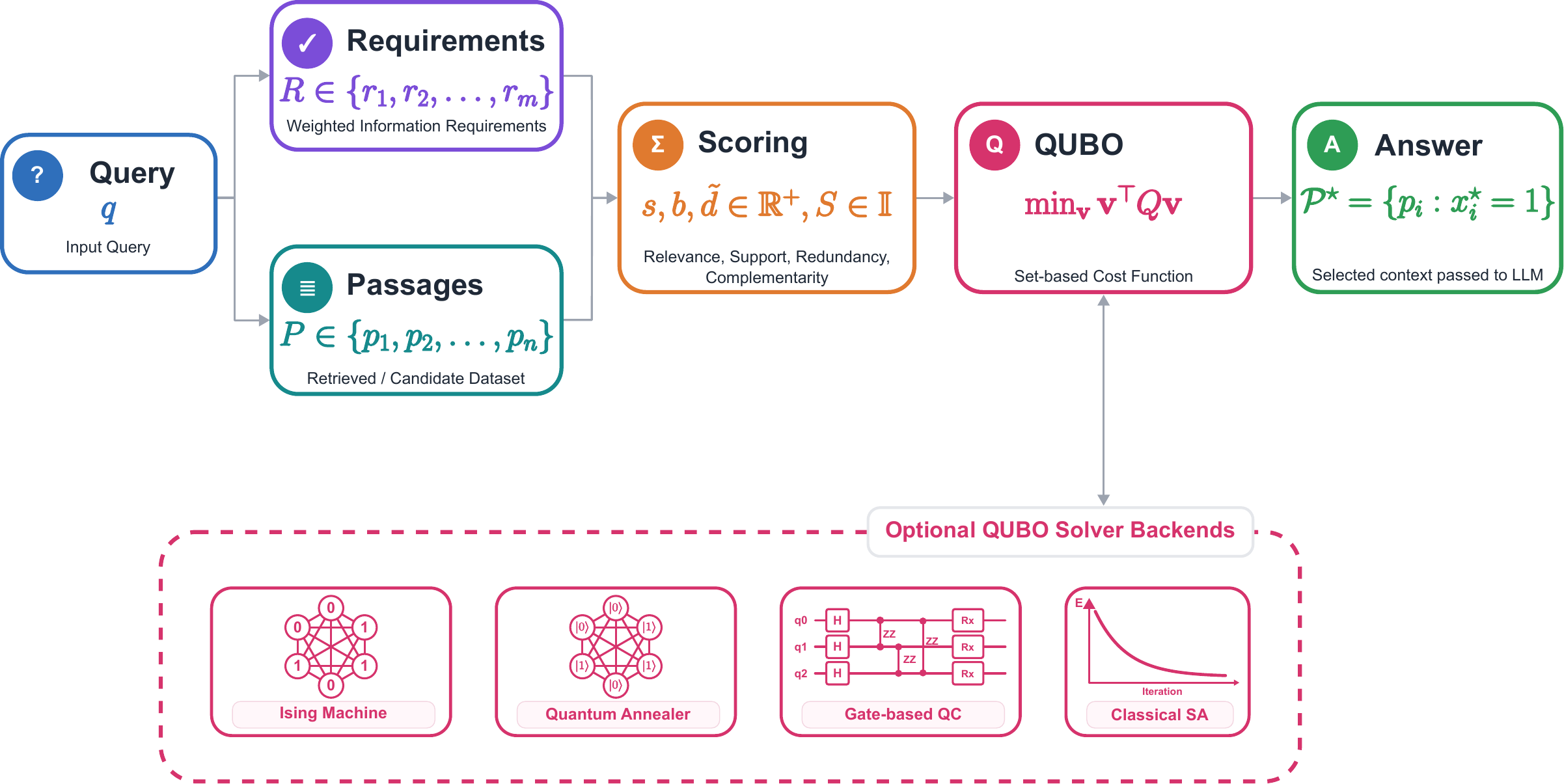}
\caption{Overview of the proposed pipeline. Given an input query, the pipeline first derives query-specific information requirements and constructs a candidate passage pool. A scoring stage defines numerical coefficients such as passage relevance \(s_i\), requirement support \(S_{ji}\), redundancy \(\tilde{d}_{ik}\), and complementarity \(b_{ik}\). These coefficients define a QUBO objective whose low-energy solution determines the selected passage subset. The resulting binary solution identifies the optimized context, which is then passed to a language model for answer generation.}
\label{fig:pipeline_workflow}
\end{figure*}

For an input query, the proposed pipeline proceeds in five stages. First, the query is decomposed into information requirements representing the evidence conditions needed to answer it. Second, a candidate passage pool is obtained from a retriever, a dataset-provided context source, or a hybrid lexical--semantic retrieval mechanism. Third, a scoring stage assigns coefficients describing passage relevance, requirement support, pairwise redundancy, and pairwise complementarity. Fourth, these coefficients parameterize a QUBO objective whose solution identifies the selected passage subset. Finally, the selected passages are concatenated with the original query and passed to a language model for answer generation. This design inserts a modular context-optimization layer between retrieval and generation: the retriever produces a broad candidate pool, the QUBO solver performs the final set-selection decision, and the downstream language model generates the answer from the optimized context. Fig.~\ref{fig:pipeline_workflow} summarizes this workflow.

\subsection{QUBO Formulation for RAG Context Selection}
\label{sec:qubo_formulation}

We formulate the final context-selection stage as a binary optimization problem. Given an input query \(q\), the objective is to select a compact passage subset that provides useful and non-redundant evidence for answer generation under a limited context budget.

Let \(\mathcal{P}=\{p_1,\dots,p_N\}\) denote the candidate passages for the query, and let \(\mathcal{R}=\{r_1,\dots,r_M\}\) denote the generated information requirements. Each requirement \(r_j\) is assigned a nonnegative importance weight \(w_j\), and each passage \(p_i\) is assigned a relevance score \(s_i\). We introduce a binary selection variable  \(x_i\in\{0,1\}\), where \(x_i=1\) indicates that passage \(p_i\) is selected for the final context and \(x_i=0\) otherwise.

The scoring stage also assigns a requirement-support strength \(S_{ji}\in\{0,1,2\}\), where \(S_{ji}=0\) indicates that passage \(p_i\) does not support requirement \(r_j\), \(S_{ji}=1\) indicates partial, bridge, or contextual support, and \(S_{ji}=2\) indicates direct support. To distinguish partial evidence from direct evidence, we define
\begin{equation}
A^{\mathrm{P}}_{ji} = \mathbbm{1}[S_{ji}\ge 1], \qquad A^{\mathrm{D}}_{ji} = \mathbbm{1}[S_{ji}=2].
\end{equation}
Here, \(A^{\mathrm{P}}_{ji}\) indicates that passage \(p_i\) provides at least partial support for requirement \(r_j\), while \(A^{\mathrm{D}}_{ji}\) indicates that the passage provides direct support.

For each requirement, we introduce two binary coverage variables
\begin{equation}
y^{\mathrm{P}}_j\in\{0,1\}, \qquad y^{\mathrm{D}}_j\in\{0,1\}.
\end{equation}
The variable \(y^{\mathrm{P}}_j=1\) indicates that requirement \(r_j\) is partially covered by at least one selected passage, while \(y^{\mathrm{D}}_j=1\) indicates that requirement \(r_j\) is directly covered by at least one selected passage. These variables are constrained by
\begin{equation}
\label{eq:setr_partial_coverage_constraint}
y^{\mathrm{P}}_j
\le
\sum_{i=1}^{N} A^{\mathrm{P}}_{ji} x_i,
\quad
y^{\mathrm{D}}_j
\le
\sum_{i=1}^{N} A^{\mathrm{D}}_{ji} x_i,
\end{equation}
We also impose the consistency constraint \(y^{\mathrm{D}}_j \le y^{\mathrm{P}}_j\), so that direct coverage cannot be activated unless partial coverage is also active. This direct/partial separation prevents multiple weak or bridge passages from being treated as equivalent to direct evidence.

To discourage selecting passages that do not support any generated requirement, we define
\begin{equation}
u_i=
\begin{cases}
1, & S_{ji}=0 \quad \forall j\in\{1,\dots,M\},\\
0, & \text{otherwise}.
\end{cases}
\end{equation}
Thus, \(u_i=1\) identifies passages with no verified requirement support.

The redundancy penalty is modulated by requirement-support overlap. Let \(\mathcal{S}_i=\{r_j\in\mathcal{R}:S_{ji}>0\}\) denote the set of requirements supported by passage \(p_i\). Given a base redundancy coefficient \(d_{ik}\) between passages \(p_i\) and \(p_k\), we define the support-aware redundancy coefficient \(\tilde{d}_{ik}=d_{ik}\rho_{ik}\), where
\begin{equation}
\label{eq:setr_rho_support_aware}
\rho_{ik}
=
\begin{cases}
J(\mathcal{S}_i,\mathcal{S}_k),
& \mathcal{S}_i \cap \mathcal{S}_k \ne \emptyset,\\
\rho_{\mathrm{diff}},
& \mathcal{S}_i \cap \mathcal{S}_k = \emptyset,
\end{cases}
\end{equation}
with \(0\le \rho_{\mathrm{diff}}<1\), and
\begin{equation}
J(\mathcal{S}_i,\mathcal{S}_k)
=
\frac{
|\mathcal{S}_i\cap\mathcal{S}_k|
}{
|\mathcal{S}_i\cup\mathcal{S}_k|
}.
\end{equation}
Passages supporting overlapping requirements therefore receive stronger redundancy penalties, while passages supporting different requirements are penalized less strongly.

We also define a pairwise complementarity coefficient \(b_{ik}\ge 0\), which rewards selecting pairs of passages that support different or complementary requirements. This term encourages the selected set to cover multiple information needs instead of repeatedly selecting passages that provide similar evidence.

For compactness, let \(\sum_i\), \(\sum_j\), and \(\sum_{i<k}\) denote sums over \(i=1,\dots,N\), \(j=1,\dots,M\), and \(1\le i<k\le N\), respectively. The constrained selection problem can be written as
\begin{equation}
\label{eq:setr_aligned_qcbo_max}
\begin{split}
    \max_{\mathbf{x},\mathbf{y}^{\mathrm{P}},\mathbf{y}^{\mathrm{D}}} 
    & \sum_i \left( \alpha s_i + \eta \sum_j w_j S_{ji} - \kappa u_i - \mu \right)x_i \\
    & + \sum_j w_j \left( \gamma_{\mathrm{P}} y^{\mathrm{P}}_j + \gamma_{\mathrm{D}} y^{\mathrm{D}}_j \right) \\
    & + \sum_{i<k} \left( \delta b_{ik} - \beta \tilde{d}_{ik} \right)x_i x_k, \\
    \text{s.t.} \hspace{0.1cm} 
    & y^{\mathrm{P}}_j \le \sum_i A^{\mathrm{P}}_{ji}x_i, \hspace{0.1cm} 
      y^{\mathrm{D}}_j \le \sum_i A^{\mathrm{D}}_{ji}x_i, \\
    & y^{\mathrm{D}}_j \le y^{\mathrm{P}}_j, 
      \hspace{0.1cm} \sum_i x_i \le K .
\end{split}
\end{equation}
where the first three constraints hold for all \(j=1,\dots,M\), \(x_i\in\{0,1\}\) for all \(i=1,\dots,N\), and \(y^{\mathrm{P}}_j,y^{\mathrm{D}}_j\in\{0,1\}\) for all \(j=1,\dots,M\).

The first line of the objective collects passage-level terms: relevance reward, support-strength reward, unsupported-passage penalty, and size penalty. The second line rewards partial and direct requirement coverage. The third line contains pairwise set-level interactions, rewarding complementary passage pairs and penalizing redundant passage pairs. The coefficients \(\alpha,\eta,\gamma_{\mathrm{P}},\gamma_{\mathrm{D}},\delta,\beta,\kappa,\mu\ge 0\) control the relative strength of these rewards and penalties. For minimization-based QUBO solvers, the equivalent constrained minimization problem is obtained by replacing the objective in \eqref{eq:setr_aligned_qcbo_max} with its additive inverse, while leaving the constraint set unchanged.

Given the binary variable domains defined above, the remaining feasibility conditions in \eqref{eq:setr_aligned_qcbo_max} are all inequality constraints. We convert these inequalities to equalities using nonnegative slack variables:
\begin{equation}
\label{eq:inequality_to_equality}
\begin{aligned}
\sum_i x_i + q_K - K &= 0,\\
y^{\mathrm{P}}_j - \sum_i A^{\mathrm{P}}_{ji}x_i + v^{\mathrm{P}}_j & = 0,\\
y^{\mathrm{D}}_j - \sum_i A^{\mathrm{D}}_{ji}x_i + v^{\mathrm{D}}_j & = 0,\\
y^{\mathrm{D}}_j-y^{\mathrm{P}}_j+m_j & = 0.
\end{aligned}
\end{equation}
The last three equalities hold for all \(j=1,\dots,M\). The slack variables \(q_K\), \(v^{\mathrm{P}}_j\), \(v^{\mathrm{D}}_j\), and \(m_j\) are nonnegative integers; in the implemented QUBO, each is represented using a binary expansion. The corresponding quadratic penalty is
\begin{equation}
\label{eq:constraint_penalty_compact}
\begin{aligned}
\mathcal{P} = & \lambda_K \left( \sum_i x_i + q_K - K \right)^2 \\
& + \lambda_{\mathrm{P}} \sum_j \left( y^{\mathrm{P}}_j - \sum_i A^{\mathrm{P}}_{ji}x_i + v^{\mathrm{P}}_j \right)^2\\
& + \lambda_{\mathrm{D}} \sum_j \left( y^{\mathrm{D}}_j - \sum_i A^{\mathrm{D}}_{ji}x_i + v^{\mathrm{D}}_j \right)^2\\
& + \lambda_{\mathrm{H}} \sum_j \left( y^{\mathrm{D}}_j - y^{\mathrm{P}}_j+m_j \right)^2 .
\end{aligned}
\end{equation}

Let \(F(\mathbf{x},\mathbf{y}^{\mathrm{P}},\mathbf{y}^{\mathrm{D}})\) denote the maximization objective in \eqref{eq:setr_aligned_qcbo_max}. The final unconstrained QUBO objective is then
\begin{equation}
\label{eq:setr_aligned_qubo}
\min_{\mathbf{x},\mathbf{y}^{\mathrm{P}},\mathbf{y}^{\mathrm{D}},\mathbf{z}}
\quad
-F(\mathbf{x},\mathbf{y}^{\mathrm{P}},\mathbf{y}^{\mathrm{D}})
+
\mathcal{P},
\end{equation}
where \(\mathbf{z}\) denotes the binary variables used to encode the slack variables, and \(\lambda_K,\lambda_{\mathrm{P}},\lambda_{\mathrm{D}},\lambda_{\mathrm{H}}>0\) are penalty weights.

The coefficients used in the objective, including relevance scores, requirement weights, support labels, redundancy coefficients, complementarity coefficients, and unsupported-passage indicators, are fixed before optimization. After the slack variables in \eqref{eq:inequality_to_equality} are encoded using binary variables, each residual in \eqref{eq:constraint_penalty_compact} is linear in the full binary variable vector. Therefore, the squared penalties introduce only linear and pairwise quadratic terms, using \(z^2=z\) for binary variables. Consequently, the unconstrained objective in \eqref{eq:setr_aligned_qubo} can be written in standard QUBO matrix form as
\begin{equation}
\label{eq:final_qubo_matrix}
\min_{\mathbf{v}\in\{0,1\}^{n_{\mathrm{Q}}}} \mathbf{v}^{\top}Q\mathbf{v},
\end{equation}
where \(\mathbf{v}\) concatenates the passage-selection variables, coverage variables, and slack-encoding variables; \(Q\) is the resulting QUBO coefficient matrix; and \(n_{\mathrm{Q}}\) is the total number of binary variables. 


\subsection{Answer Generation}
\label{sec:solver_output_context}

Let \(\mathbf{v}^{\star}\) denote the binary solution returned by the QUBO solver. The selected passage subset is recovered from the passage-selection variables as
\begin{equation}
\mathcal{P}^{\star} = \{p_i\in\mathcal{P}:x_i^{\star}=1\}.
\end{equation}
The selected passages are then assembled into the final context and concatenated with the original query to form the answer-generation prompt.

\section{Results}
\label{sec:results}

We evaluate the proposed QUBO selector from three perspectives. First, we perform a development-set sweep over QUBO configurations to select a fixed operating point. Second, we compare answer-generation performance against SetR-style LLM set selectors and non-LLM selection baselines. Third, we analyze the selected evidence sets directly through ablation experiments. Unless otherwise stated, we use an integer-quantized QUBO representation in which real-valued scoring coefficients are scaled and rounded to integer values before solving. This quantization controls coefficient precision and dynamic range, which is important for finite-precision QUBO and Ising solvers, including hardware-oriented implementations with limited coefficient resolution.

\subsection{Experimental Setup}
\label{sec:experimental_setup}

We evaluate the proposed selector on the HotpotQA distractor setting~\cite{Yang2018}, where each example contains a multi-hop question, a reference answer, and context passages containing both supporting and distractor evidence. To make evidence selection explicit, each titled context is flattened into sentence-level candidate passages, and all compared methods operate on the same candidate pool \(\mathcal{P}=\{p_i\}_{i=1}^{N}\). All LLM-related components use the same local model, \texttt{Ollama 3.6:31b}, including information-requirement generation, passage-support labeling, LLM-based selector baselines, and final answer generation. This setup keeps the candidate pool and generator fixed across methods, so differences in performance primarily reflect the context-selection strategy. In the proposed pipeline, the LLM provides semantic measurements used to construct the QUBO, while the final passage subset is selected by the QUBO solver. This modular design separates semantic preprocessing, combinatorial context selection, and answer generation, allowing each component to be replaced independently.

\subsection{Development-Set Configuration Sweep}
\label{sec:configuration_sweep}

We first evaluate a development-set sweep over QUBO configurations. All configurations use the same base objective weights \(\alpha=10\), \(\beta=10\), \(\gamma_{\mathrm{P}}=4\), \(\gamma_{\mathrm{D}}=10\), \(\delta=8\), support reward \(10\), unsupported-passage penalty \(20\), and \(\rho_{\mathrm{diff}}=0.25\). The sweep varies constraint penalties, passage budget, and compactness penalty. Table~\ref{tab:configuration_sweep} summarizes the resulting answer-level and set-level behavior.

\begin{table}[t]
\centering
\scriptsize
\setlength{\tabcolsep}{2.8pt}
\begin{tabular}{llcccc}
\hline
\textbf{Cfg.} & \textbf{Changed setting} & \textbf{EM} & \textbf{F1} & \textbf{Cov.} & \textbf{\#Sel.} \\
\hline
\(C_0\) & baseline & 0.6000 & 0.7151 & 0.9883 & 4.39 \\
\(C_1\) & weak constraints & \textbf{0.6200} & \textbf{0.7485} & 0.9883 & 4.61 \\
\(C_2\) & strong constraints & 0.6000 & 0.7218 & 0.9833 & 4.40 \\
\(C_3\) & \(K=3\) & 0.6000 & 0.7051 & 0.9700 & 3.05 \\
\(C_4\) & \(K=4\) & 0.6000 & 0.7151 & 0.9700 & 3.70 \\
\(C_5\) & \(K=6\) & 0.6100 & 0.7318 & 0.9883 & 5.01 \\
\(C_6\) & low cov. pen. & 0.6000 & 0.7151 & 0.9883 & 4.37 \\
\(C_7\) & high cov. pen. & 0.5900 & 0.7185 & 0.9783 & 4.38 \\
\(C_8\) & very high cov. pen. & 0.6100 & 0.7418 & 0.9883 & 4.51 \\
\(C_9\) & low budget pen. & 0.6100 & 0.7465 & 0.9883 & 4.78 \\
\(C_{10}\) & high budget pen. & 0.5800 & 0.7154 & 0.9733 & 4.54 \\
\(C_{11}\) & low size pen. & 0.6000 & 0.7218 & 0.9883 & 4.37 \\
\(C_{12}\) & high size pen. & 0.6000 & 0.7151 & 0.9883 & 4.40 \\
\hline
\end{tabular}
\caption{Development-set sweep over QUBO configurations using simulated annealing. Cov. denotes average requirement coverage and \#Sel. denotes the average number of selected passages.}
\label{tab:configuration_sweep}
\end{table}

The QUBO selector is relatively stable across nearby settings: most configurations maintain high requirement coverage while varying in compactness and downstream answer quality. The best answer-level performance is obtained by \(C_1\), which weakens the budget and coverage penalties relative to the baseline while preserving high coverage. Reducing the budget to \(K=3\) produces more compact contexts but lowers F1, while increasing the budget to \(K=6\) selects more passages without improving over \(C_1\). Very large budget penalties also reduce performance, suggesting that overly rigid constraint enforcement can interfere with the selector's ability to balance coverage, support, and compactness.

\subsection{Comparison with Prior LLM-Based Set Selection}
\label{sec:setr_comparison}

We next compare the proposed QUBO selector against SetR-style LLM set-selection baselines inspired by SetR~\cite{Lee2025}. Since the official SetR model is not publicly available to us, we implement prompt-based SetR-style selectors using the same local \texttt{qwen3.6:35b} model used elsewhere in the pipeline. \textbf{SetR} denotes the SetR-aligned baseline in which the model is given the precomputed information requirements and selects at most \(K\) passages from the same candidate pool. And, \textbf{SetRO} denotes the original-prompt-style baseline in which the model infers information requirements from the question and candidate passages before selecting the final evidence set. Thus, SetR tests LLM-based selection with the same requirement structure used by the QUBO formulation, while SetRO tests a closer SetR-style prompting setup where requirements are inferred inside the LLM prompt.

\begin{table}[t]
\centering
\small
\setlength{\tabcolsep}{4pt}
\begin{tabular}{lcccc}
\hline
\textbf{Method} & \textbf{EM} & \textbf{F1} & \textbf{Cov.} & \textbf{\#Sel.} \\
\hline
QUBO selector & 0.6500 & 0.7866 & \textbf{0.9893} & 3.862 \\
SetR & 0.6540 & 0.7930 & 0.9847 & 2.320 \\
SetRO & \textbf{0.6580} & \textbf{0.8026} & 0.9845 & \textbf{2.306} \\
\hline
\end{tabular}
\caption{Comparison with SetR-style LLM set selectors on 500 HotpotQA examples. All methods use the same candidate pool and answer-generation model.}
\label{tab:setr_answer_comparison}
\end{table}

Table~\ref{tab:setr_answer_comparison} shows that the QUBO selector is competitive with direct LLM-based set selection. SetR and SetRO obtain slightly higher answer-level scores, but the margin is small: the QUBO selector is within \(0.004\) EM and \(0.0064\) F1 of SetR, and within \(0.008\) EM and \(0.0160\) F1 of SetRO. At the same time, the QUBO selector achieves the highest requirement coverage, indicating that it tends to select contexts that more explicitly satisfy the generated information requirements. This supports the intended role of the QUBO layer: it replaces direct LLM set selection with explicit subset optimization while preserving competitive downstream answer quality.

\subsection{Comparison with Non-LLM Selectors}
\label{sec:non_llm_results}

We also compare against non-LLM selection baselines: relevance top-\(k\), maximal marginal relevance (MMR), hybrid lexical--semantic ranking, BM25, greedy coverage, and random selection~\cite{Robertson2009, Carbonell1998, karpukhin2020, ma2022}. These methods provide reference points for relevance-only selection, diversity-aware selection, lexical retrieval, coverage-driven selection, and chance-level selection.

\begin{table}[t]
\centering
\small
\setlength{\tabcolsep}{4pt}
\begin{tabular}{lcc}
\hline
\textbf{Method} & \textbf{EM} & \textbf{F1} \\
\hline
Relevance top-\(k\) & \textbf{0.6500} & \textbf{0.7718} \\
MMR-Relevance & \textbf{0.6500} & 0.7651 \\
QUBO selector & \textbf{0.6500} & 0.7585 \\
Greedy Coverage Compact & 0.6100 & 0.7296 \\
Hybrid Rel.+BM25 top-\(k\) & 0.5900 & 0.7320 \\
MMR-BM25 & 0.4100 & 0.5061 \\
BM25 top-\(k\) & 0.4000 & 0.4941 \\
Random & 0.0700 & 0.0786 \\
\hline
\end{tabular}
\caption{Answer-level comparison with non-LLM selection baselines on 100 HotpotQA examples.}
\label{tab:non_llm_baselines}
\end{table}

Table~\ref{tab:non_llm_baselines} shows that the QUBO selector matches the best exact-match score among non-LLM baselines and remains close to relevance top-\(k\) and MMR-Relevance in F1. The QUBO selector substantially outperforms BM25-based and random selection methods, confirming that the selected contexts are not simply the result of lexical overlap or chance. Although relevance top-\(k\) obtains the highest F1 in this subset, it does not explicitly optimize requirement coverage, redundancy, complementarity, or compactness. The QUBO formulation therefore provides a structured alternative to relevance-only context construction while maintaining competitive answer-level performance.

\subsection{Ablation Study}
\label{sec:ablation_study}

We ablate the main components of the QUBO objective to examine how different terms affect both downstream answer generation and the structure of the selected evidence set. The answer-level ablation measures whether the selected context enables the generator to produce the correct answer, while the set-level ablation isolates the behavior of the selector itself.

\subsubsection{Answer-Level Ablation}
\label{sec:answer_ablation}

Table~\ref{tab:answer_ablation} reports exact match and token-level F1 for the full QUBO objective and its ablated variants.

\begin{table}[t]
\centering
\small
\setlength{\tabcolsep}{4pt}
\begin{tabular}{lcc}
\hline
\textbf{Variant} & \textbf{EM} & \textbf{F1} \\
\hline
Rel.+Coverage only & \textbf{0.6600} & \textbf{0.7751} \\
No Complementarity & \textbf{0.6600} & 0.7685 \\
No Redundancy & \textbf{0.6600} & 0.7685 \\
No Size Penalty & \textbf{0.6600} & 0.7685 \\
No Unsupported Penalty & \textbf{0.6600} & 0.7685 \\
Full QUBO & 0.6500 & 0.7585 \\
No Support Strength & 0.6500 & 0.7518 \\
\hline
\end{tabular}
\caption{Answer-level ablation of QUBO objective components on 100 HotpotQA examples.}
\label{tab:answer_ablation}
\end{table}

The results show that relevance and requirement coverage are the strongest contributors to answer-level performance on this subset. The relevance and coverage only variant obtains the highest F1, while removing support strength produces the lowest F1 among the tested variants. The full QUBO objective is slightly lower in answer F1 than some ablations, indicating that optimizing for richer set-level structure does not always translate directly into the best short-answer score. This indicates that the objective terms that improve evidence-set structure do not always yield proportional gains in exact-match or token-F1, which are indirect measures of selector quality.

\subsubsection{Set-Level Ablation}
\label{sec:set_level_ablation}

To isolate the behavior of the selector itself, Table~\ref{tab:set_level_ablation} evaluates the selected evidence sets directly using weighted coverage, capped support, marginal utility, average relevance, and selected-context size.

\begin{table}[t]
\centering
\scriptsize
\setlength{\tabcolsep}{3pt}
\begin{tabular}{lccccc}
\hline
\textbf{Variant} & \textbf{WC} & \textbf{CS} & \textbf{MU} & \textbf{Rel.} & \textbf{\#Sel.} \\
\hline
Full QUBO & 0.9852 & 0.9693 & \textbf{0.5056} & 7.4156 & 3.5667 \\
No Redundancy & 0.9852 & 0.9693 & 0.5039 & 7.3856 & 3.6000 \\
No Complementarity & 0.9852 & 0.9693 & \textbf{0.5056} & \textbf{7.4222} & \textbf{3.5333} \\
No Support Strength & 0.9852 & 0.9693 & \textbf{0.5056} & 7.4139 & \textbf{3.5333} \\
No Size Penalty & 0.9852 & 0.9693 & \textbf{0.5056} & 7.4156 & 3.5667 \\
No Unsupported Penalty & 0.9852 & 0.9693 & 0.4417 & 6.9017 & 3.8333 \\
Rel.+Coverage only & 0.9852 & 0.9693 & 0.4194 & 6.5778 & 3.9667 \\
\hline
\end{tabular}
\caption{Set-level ablation results. WC denotes weighted coverage, CS denotes capped support, MU denotes marginal utility fraction, Rel. denotes average relevance, and \#Sel. denotes the average number of selected passages.}
\label{tab:set_level_ablation}
\end{table}

The set-level results show that all variants maintain the same weighted coverage and capped support, indicating that the requirement-coverage constraints are robust across these ablations. The differences instead appear in compactness, marginal utility, and average relevance. Removing the unsupported-passage penalty or using only relevance and coverage increases the number of selected passages while reducing marginal utility and average relevance, suggesting that these variants include more low-utility context even when they preserve coverage. In contrast, the full QUBO maintains high coverage while selecting a more compact and higher-utility evidence set. Together, the answer-level and set-level ablations show that simple relevance-and-coverage objectives can be strong for answer generation, while the full QUBO better controls the structure and compactness of the selected evidence set.

\section{Conclusion}
\label{sec:conclusion}

We presented a QUBO-based formulation for requirement-aware context selection in retrieval-augmented question answering. Instead of selecting passages solely by independent relevance scores or asking an LLM to directly choose the final context, the proposed method casts evidence selection as a binary subset-optimization problem. Candidate passages are represented by binary variables, and the objective combines passage relevance, requirement coverage, support strength, redundancy, complementarity, compactness, and budget constraints within a single solver-compatible formulation. The selected passages are then passed to a downstream language model for answer generation, separating combinatorial evidence selection from semantic answer synthesis.

Experiments on HotpotQA show that the QUBO selector achieves competitive answer-level performance relative to SetR-style LLM selectors and non-LLM baselines while providing a transparent optimization objective for the intermediate context-selection step. The QUBO selector is within a small margin of SetR-style LLM baselines in exact match and token-level F1, and it achieves strong requirement coverage across experiments. Ablation results further show that relevance and requirement coverage are the dominant contributors to downstream answer accuracy, while additional terms such as unsupported-passage penalties, redundancy, complementarity, and compactness shape the quality and structure of the selected evidence set.

These results suggest that multi-hop RAG context construction can be treated as a structured discrete optimization problem rather than only as a ranking or prompting problem. The formulation is modular: semantic preprocessing, QUBO solving, and answer generation can be replaced independently. This opens a path toward retrieval pipelines in which LLMs are reserved primarily for semantic measurement and answer generation, while the final combinatorial selection step is handled by classical, quantum-inspired, or hardware-oriented QUBO solvers. Future work includes evaluating larger and more diverse QA benchmarks, reducing reliance on LLM-based semantic preprocessing, learning the QUBO coefficients from data, and studying the behavior of the formulation on specialized Ising/QUBO hardware.

\section{Acknowledgment}
The authors gratefully acknowledge Prof. Xifeng Yan for his thoughtful discussions, valuable feedback, and insightful comments, which helped improve the presentation and positioning of this work.


\bibliography{custom}




\end{document}